\pdfoutput=1
\documentclass[10pt,journal,compsoc]{IEEEtran} %

\usepackage{amsmath,amssymb,amsfonts}
\usepackage{graphicx}
\usepackage{caption}
\usepackage{subcaption}
\usepackage{makecell}
\usepackage{booktabs}
\usepackage{siunitx}
\usepackage{adjustbox} %
\usepackage{array}
\usepackage{multirow}
\usepackage{rotating}
\usepackage{etoolbox}
\robustify\bfseries

\usepackage{url}
\usepackage{color}
\usepackage{balance} %

\def\MYTITLE{A Fast Geometric Regularizer to Mitigate Event Collapse in the Contrast Maximization Framework} %

\usepackage[pagebackref=false,breaklinks=true,colorlinks,bookmarks=true,bookmarksnumbered=true]{hyperref}
\hypersetup{
  pdftitle={\MYTITLE},
  pdfsubject={Computer Vision, Robotics, Artificial Intelligence},
  pdfauthor={Shintaro Shiba, Yoshimitsu Aoki, Guillermo Gallego},
  pdfkeywords={Event Cameras, Intelligent Systems, Asynchronous Sensor, High Temporal Resolution}
}

\def\tref{t_\text{ref}} %
\def\pol{p} %

\def\cE{\mathcal{E}} %
\def\numEvents{N_e} %
\def\numPixels{N_p} %
\def\Warp{\mathbf{W}}
\def\btheta{\boldsymbol{\theta}} %

\def\bx{\mathbf{x}}
\def\bparams{\btheta}
\def\Rot{\mathtt{R}}
\def\br{\mathbf{r}} %

\def\pol{p}
\def\velflow{\mathbf{v}}
\def\linvel{\mathbf{V}} %
\def\angvel{\boldsymbol{\omega}} %
\def\bphi{\boldsymbol{\phi}}

\def\cN{\mathcal{N}} %
\def\bmu{\boldsymbol{\mu}} %

\def\flow{\mathbf{f}}
\def\cD{\mathcal{D}} %
\def\cA{\mathcal{A}} %

\def\regularizer{\mathcal{R}}
\def\depthx{Z(\bx)} %

\def\IWE{I} %
\def\bzero{\mathbf{0}}
\def\mId{\mathtt{Id}} %
\def\variance{\operatorname{Var}}
\def\mJ{\mathtt{J}}
\def\bvec{\mathbf{b}}
\def\mA{\mathtt{A}}

\def\mH{\mathtt{H}} %
\def\be{\mathbf{e}}
\def\bX{\mathbf{X}}

\def\bgamma{\boldsymbol{\gamma}} %

\def\mJdelta{\mJ_{t,t+\Delta t}}
\def\mHdelta{\mH_{t,t+\Delta t}}
\def\mJderivZero{\left.\frac{d |\mJdelta|}{d \Delta t}\right|_{\Delta t = 0}}

\usepackage{orcidlink}

\usepackage[capitalize]{cleveref}
\newif\ifaisy
\aisytrue %
\ifaisy
\crefname{section}{Section}{Sections}
\crefname{table}{Table}{Tables}
\crefname{figure}{Figure}{Figures}
\else 
\crefname{section}{Sec.}{Secs.}
\crefname{table}{Tab.}{Tabs.} 
\crefname{figure}{Fig.}{Figs.}
\fi
\Crefname{section}{Section}{Sections}
\Crefname{table}{Table}{Tables}
\Crefname{figure}{Figure}{Figures}

\usepackage{cite}

\newcommand{\unumr}[2]{\multicolumn{1}{r}{\underline{\tablenum[table-format={#1}]{#2}}}}  %
\newcommand{\bnum}[1]{\bfseries #1}

\newcommand{\anote}[1]{\multicolumn{1}{r}{{$^*$}{#1}}}  %
\newcommand{\novalue}{{\textendash}}

\definecolor{light-gray}{gray}{0.6}
\newcommand\gframe[1]{{\color{light-gray}\frame{#1}}}

\makeatletter
\long\def\@IEEEtitleabstractindextextbox#1{\parbox{0.922\textwidth}{#1}}
\makeatother

\usepackage[absolute]{textpos}

\begin{document}

\definecolor{somegray}{gray}{0.66}
\newcommand{\darkgrayed}[1]{\textcolor{somegray}{#1}}
\begin{textblock}{11}(2.5, 0.45)
\begin{center}
\darkgrayed{This journal paper has been accepted for publication at Advanced Intelligent Systems, 2022.}
\end{center}
\end{textblock}

\title{\MYTITLE}

\author{Shintaro Shiba$^{1,2}$\orcidlink{0000-0001-6053-2285}, Yoshimitsu Aoki$^{1}$, Guillermo Gallego$^{2,3}$\orcidlink{0000-0002-2672-9241}%
\IEEEcompsocitemizethanks{\IEEEcompsocthanksitem 
$^1$ Department of Electronics and Electrical Engineering, Faculty of Science and Technology, Keio University, Kanagawa, Japan. 
$^2$ Department of Electrical Engineering and Computer Science, Technische Universit\"at Berlin, Berlin, Germany. 
$^3$ Einstein Center Digital Future and Science of Intelligence Excellence Cluster, Berlin, Germany.
\IEEEcompsocthanksitem Preprint of paper accepted at Advanced Intelligent Systems, 2022.
\hfil\break doi: 10.1002/aisy.202200251.
}%
}

\IEEEtitleabstractindextext{%
\begin{abstract}
Event cameras are emerging vision sensors and their advantages are suitable for various applications such as autonomous robots.
Contrast maximization (CMax), which provides state-of-the-art accuracy on motion estimation using events,
may suffer from an overfitting problem called event collapse.
Prior works are computationally expensive or cannot alleviate the overfitting, which undermines the benefits of the CMax framework.
We propose a novel, computationally efficient regularizer based on geometric principles to mitigate event collapse.
The experiments show that the proposed regularizer achieves state-of-the-art accuracy results, while its reduced computational complexity makes it two to four times faster than previous approaches.
To the best of our knowledge, our regularizer is the only effective solution for event collapse without trading off runtime.
We hope our work opens the door for future applications that unlocks the advantages of event cameras.

\end{abstract}
}
\maketitle

\section*{Multimedia Material}
Project page: \url{https://github.com/tub-rip/event_collapse}

\section{Introduction}
Event cameras \cite{Lichtsteiner08ssc,Finateu20isscc} are an emerging technology attracting considerable attention because of their advantages, such as low latency, high dynamic range and data/power efficiency.
These advantages are especially suitable for applications that require low-latency vision-based processing, such as autonomous driving \cite{Chen20msp,Murali22aisy}.
The neuromorphic principle of operation of event cameras makes them naturally respond to the scene dynamics.
Therefore, they have been used in motion estimation tasks 
\cite{Wang21aisy,Zhong22aisy}, such as Simultaneous Localization and Mapping \cite{Kim16eccv,Rebecq17ral,Zhu17cvpr,Zhou20tro,Rosinol18ral,Rebecq18ijcv},
optical flow estimation \cite{Benosman14tnnls,Zhu19cvpr,Paredes21neurips,Shiba22eccv},
particle image velocimetry \cite{Willert22expfl},
motion segmentation \cite{Stoffregen19iccv,Zhou21tnnls},
obstacle avoidance \cite{Falanga20scirob},
and space situational awareness~\cite{Cohen19jas}.

\begin{figure}[t]
\centering
\begin{subfigure}{.94\linewidth}
  \centering
  {\includegraphics[clip,trim={0cm 0.98cm 0cm 1.03cm},width=\linewidth]{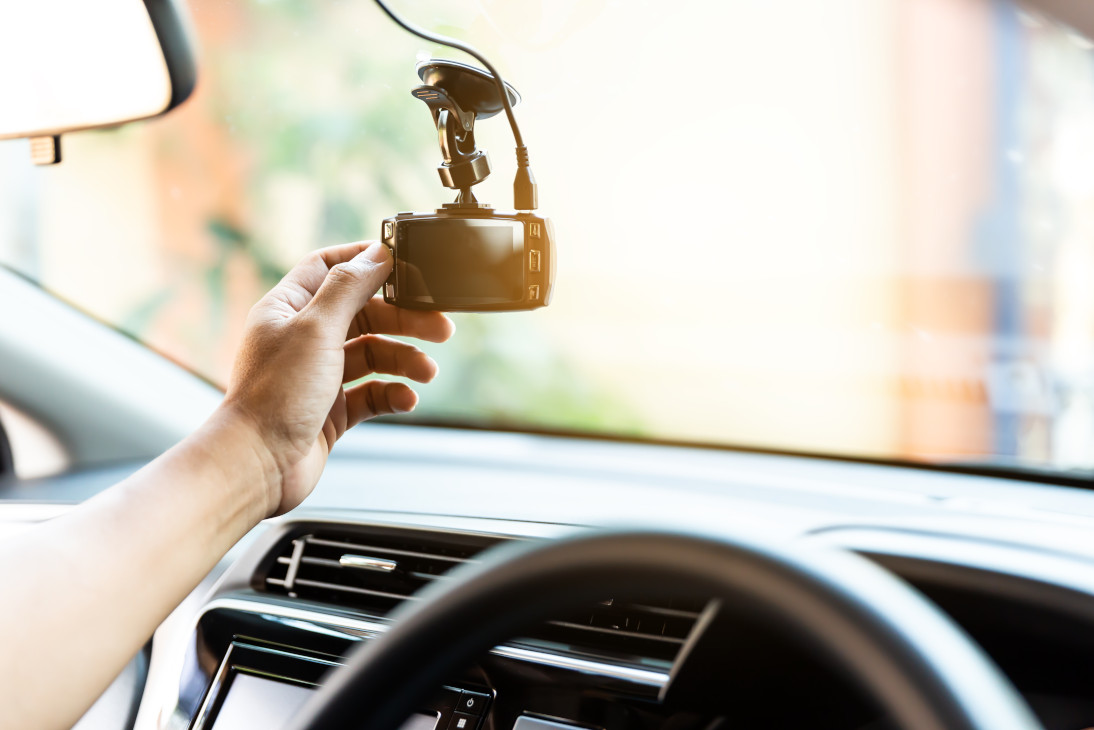}}
\end{subfigure}\\[1ex]
\begin{subfigure}{.45\linewidth}
  \centering
  \gframe{\includegraphics[width=\linewidth]{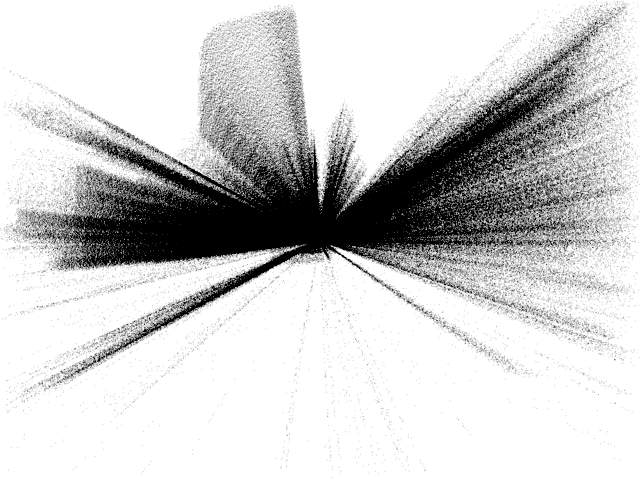}}
\end{subfigure}\;\;
\begin{subfigure}{.45\linewidth}
  \centering
  \gframe{\includegraphics[width=\linewidth]{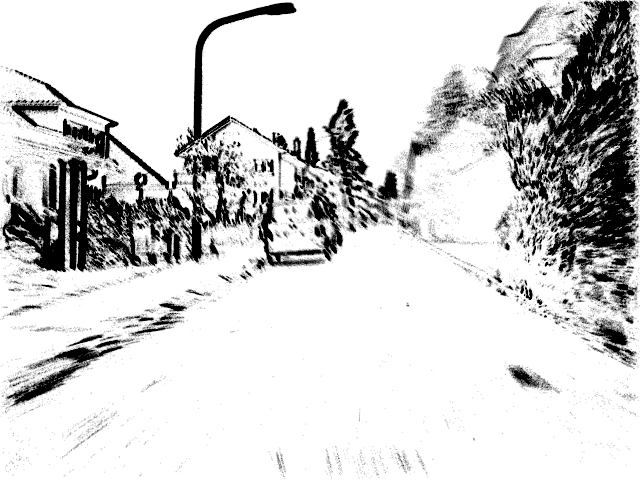}}
\end{subfigure}
\\
\caption{\emph{Sample application of event cameras.} 
Top: The advantages of event cameras are beneficial for robotics applications, such as autonomous driving. 
Bottom: The proposed regularizer discourages event collapse (left), and reveals sharp edges of the scene in a computationally efficient manner (right).
(Top image licensed from Stock Photo ID 1102269152).
}
\label{fig:eyecatcher}
\end{figure}

Contrast maximization (CMax) \cite{Gallego18cvpr} is a powerful event processing framework that achieves state-of-the-art accuracy in various motion estimation tasks.
On the other hand, it may suffer from an overfitting problem called \emph{event collapse}, 
where the optimizer converges to an undesired global optimum \cite{Shiba22sensors}.
Prior works have tackled event collapse in several ways, 
such as whitening the data \cite{Nunes21pami}, 
reformulating the task (e.g., by providing additional depth information \cite{Nunes21pami}),
or adding a regularizer to improve the optimization landscape \cite{Shiba22sensors}.
However, these proposals present shortcomings: 
($i$) assuming known depth data is task-specific and requires an additional sensor such as a LiDAR or a stereo setup,
and ($ii$) the above regularizing techniques may not be effective \cite{Nunes21pami} or require considerable extra computation \cite{Shiba22sensors}.
Towards more practical application of event cameras, it is paramount to effectively alleviate event collapse in a computationally-efficient manner.

\begin{figure*}[t]
  \centering
  {\includegraphics[clip,trim={0cm 9cm 0cm 0cm},width=0.9\linewidth]{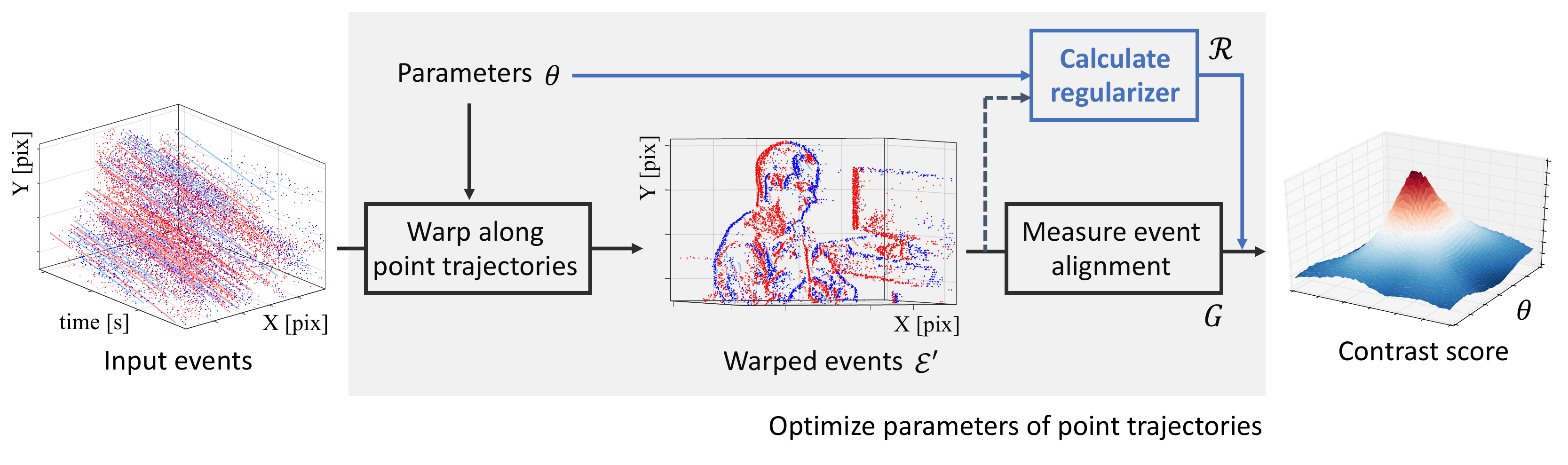}}
\caption{\emph{Method overview}. 
The proposed regularizer (blue line) is based on geometric principles and solely relies on motion parameters $\bparams$, while previous approaches (dashed line) are built from warped events \cite{Shiba22sensors}.
Adapted with permission from Ref.~\cite{Gallego19cvpr}, 2019, Gallego et al.
}
\label{fig:method:overview}
\end{figure*}

This paper proposes a novel, computationally efficient regularizer to mitigate event collapse in the CMax framework (\cref{fig:eyecatcher}).
From a theoretical point of view, the regularizer is designed based on geometric principles of motion field deformation (measuring area rate of change along point trajectories).
In contrast to previous methods, the regularizer does not depend on the event data; it only depends on the motion hypothesis (i.e., the warp). 
This is desirable because events may not be equally distributed on the space-time image domain, 
and motion hypotheses provide information even in homogeneous brightness regions (where events are scarce).
From a practical point of view, the proposed regularizer 
drastically reduces computational complexity, being two to four times faster than previous solutions, 
while achieving state-of-the-art results to mitigate event collapse.

The above contributions open the door to efficient and interpretable regularizers for motion estimation problems with geometrically-meaningful parametrizations.

\section{Related Work}

Motion estimation from event data is an incipient research field. 
Event cameras respond to moving edges in the scene, and edges carry the essential information for motion estimation. 
Hence event cameras have numerous applications in motion-related tasks 
(ego-motion and/or scene-motion).
We refer to \cite{Gallego20pami} for a comprehensive review.
State-of-the-art approaches are based on the concept of \emph{event alignment} or motion compensation \cite{Gallego18cvpr,Gallego19cvpr,Nunes21pami,Gu21iccv}, estimating the motion parameters that associate events corresponding to the same scene edge.
In particular, Contrast Maximization (CMax) \cite{Gallego18cvpr} seeks the motion parameters that maximize the contrast (or focus score) of an image of warped events (IWE).
It has been used on various motion estimation tasks,
such as rotational motion \cite{Gallego17ral,Kim21ral,Gu21iccv},
feature flow \cite{Zhu17icra,Zhu17cvpr,Seok20wacv,Stoffregen19cvpr}, 
homographic motion \cite{Gallego18cvpr,Nunes21pami,Peng21pami},
optical flow \cite{Zhu19cvpr,Paredes21neurips,Shiba22eccv}, 
motion segmentation \cite{Mitrokhin18iros,Stoffregen19iccv,Zhou21tnnls,Parameshwara21icra} 
and 3D reconstruction \cite{Rebecq18ijcv,Ghosh22aisy}.
However, the optimizer may converge to an undesired global optimum that pushes events into too few pixels (i.e., event collapse).

Event collapse is originally analyzed in detail by \cite{Shiba22sensors}.
It may occur or not depending on the task, the space of motion hypotheses and the data.
Hence, previous works for tackling event collapse can be categorized based on tasks.
For example, optical flow estimation has a high number of degrees-of-freedom (DOFs) ($2\numPixels$, where $\numPixels$ is the number of pixels), i.e., motion parameters, and is a collapse-enabled problem.
A common approach to mitigate collapse for this task is to add a strong regularizer, such as the classical Charbonnier loss \cite{Charbonnier97tip}, to encourage smoothness of the flow \cite{Zhu19cvpr,Stoffregen19cvpr}.
Another approach consists of increasing the well-posedness of the problem by using a tile-based motion field and a multi-reference focus loss \cite{Shiba22eccv}.
However, event collapse may still appear in some parts of the image plane.

Ego-motion estimation problems, which are the main focus of this work, parametrize motion over the image space with relatively lower DOFs.
The well-posedness of the problem is affected not only by the number of DOFs but also by the geometric meaning of the motion.
By reformulating the problem to reduce the number of DOFs, \cite{Ozawa22sensors} and \cite{Peng21pami} increase the well-posedness of the task.
Whitening of warped events is proposed in \cite{Nunes21pami} to mitigate event collapse,
while \cite{Shiba22sensors} designs effective regularizers based on the divergence or the area deformation of the motion field, at the expense of increased computational cost. 
Initialization close to the solution (in the basin of attraction of the desired local optimum) can also play an important role in evading event collapse \cite{Gallego18cvpr}. 

Our work is most related to \cite{Shiba22sensors} because we focus on low-DOF motion estimation tasks
and seek a principled regularizer to gauge and penalize event collapse.
Our theoretical analysis provides a formula for homographic motions (8~DOFs), which can be particularized for: 
1~DOF (zoom-in/out motion), 2~DOFs (feature flow), 3~DOFs (rotational camera motion), 4~DOFs (planar similarity motion) and 6~DOFs (planar affine motion).
While both \cite{Shiba22sensors} and our proposal are interpretable and grounded on geometric principles of motion trajectories, the most important theoretical difference is that our regularizer does not depend on the event data. 
As a result, the regularizer drastically improves computational complexity while achieving on par or better motion estimation results on publicly-available datasets.

\section{Methodology}
\label{sec:method}

This section first reviews how an event camera works (\cref{sec:method:eventcamera}) 
as well as the regularized CMax framework (\cref{sec:method:cmaxreg}).
Then, we propose the new regularizer and explain its geometrical meaning and implications (\cref{sec:method:proposedReg}).

\subsection{Event Camera}
\label{sec:method:eventcamera}

Instead of acquiring brightness images at fixed time intervals (e.g., frames), event cameras record brightness differences asynchronously, called ``events'' \cite{Lichtsteiner08ssc,Gallego20pami}.
An event $e_k \doteq (\bx_k, t_k, \pol_{k})$ represents a brightness change,
and it is triggered as soon as the logarithmic brightness at the pixel $\bx_k\doteq (x_k, y_k)^{\top}$ exceeds a preset threshold.
Here, $t_k$ is the timestamp of the event with \si{\micro\second} resolution,
and polarity $\pol_{k} \in \{+1,-1\}$ is the sign of the brightness change.
\Cref{fig:method:overview} shows the input events (red and blue dots ($\bx_k,t_k$) in space-time, with color representing polarity).

\subsection{Regularized Contrast Maximization}
\label{sec:method:cmaxreg}

The regularized CMax framework \cite{Shiba22sensors} aims at finding the motion parameters $\bparams$ that optimize the objective function:
\begin{equation}
\label{eq:compositeObjective}
\bparams^\ast %
= \arg\!\min_{\bparams} \left(-G(\bparams) + \lambda \regularizer(\bparams)\right).
\end{equation}

Here, $G$ is the data fidelity term and $\regularizer$ is a regularizer with weight $\lambda > 0$.
The overall steps of the framework are described in \cref{fig:method:overview}, where black lines indicate the previous approaches, and blue lines indicate our proposal.

\subsubsection{Data fidelity term}
The term $G(\bparams)$ measures the alignment of the events with respect to the candidate motion $\bparams$ (\cref{fig:method:overview}, black solid line).
The original events $\cE = \{e_k\}_{k=1}^{\numEvents}$ are transformed according to the motion hypothesis into a set of warped events $\cE' = \{e'_k\}_{k=1}^{\numEvents}$:
\begin{equation}
\label{eq:warp}
e_k \doteq (\bx_k,t_k,\pol_k) \quad\stackrel{\Warp}{\mapsto}\quad
e'_k \doteq (\bx'_k,\tref,\pol_k).
\end{equation}
The warp function $\bx'_k = \Warp(\bx_k,t_k; \bparams)$ transports every event along its motion trajectory until a reference time $\tref$ is reached.
Point trajectories are parametrized by $\bparams$, which consists of motion or scene unknowns (e.g., scene depth, moving objects).

Powerful objective functions are designed based on the count of warped events \cite{Gallego19cvpr}.
The representation of event count as an image is defined by the image of warped events (IWE):
\begin{equation}
\label{eq:IWE}
\IWE(\bx;\bparams) \doteq \sum_{k=1}^{\numEvents} b_k \,\delta (\bx - \bx'_k(\bparams)).
\end{equation}

Each pixel of IWE counts how many events $e'_k$ are warped into pixel $\bx$.
The event polarity can be used by setting $b_k=\pol_k$, and not used if $b_k=1$.
The Dirac delta $\delta$ is approximated by a Gaussian: $\delta(\bx-\bmu)\approx\cN(\bx;\bmu,\epsilon^2)$, where $\epsilon=1$ pixel.

Finally, the objective function, such as the IWE variance is calculated:
\begin{equation}
\label{eq:IWEVariance}
G(\bparams) \equiv \variance\bigl(\IWE(\bx;\bparams)\bigr) 
\doteq \frac{1}{|\Omega|} \int_{\Omega} (\IWE(\bx;\bparams)-\mu_{\IWE})^2 d\bx,
\end{equation}
with mean $\mu_{\IWE} \doteq \frac{1}{|\Omega|} \int_{\Omega} \IWE(\bx;\bparams) d\bx$ and image domain $\Omega$.
The interpretation of \eqref{eq:IWEVariance} is as follows: 
the larger the IWE variance (contrast of the IWE), the better the alignment of the warped events $\cE'$.
Contrast is related to sharpness and focus \cite{Gallego19cvpr}.

\subsubsection{Previous regularizers}
The regularizer term $\regularizer$ in \eqref{eq:compositeObjective} penalizes event collapse in certain types of warps.
When event collapse happens, the warped events are accumulated into too few pixels or lines, resulting in an undesired global optimum $G$ (overfitting).
Two regularizers are proposed in \cite{Shiba22sensors} by averaging some collapse quantities attached to each event (\cref{fig:method:overview}, dashed line).
Specifically, they use the divergence of the flow $\cD (\cE,\bparams)$ and the area-based deformation of the warp $\cA (\cE,\bparams)$, which are given by:
\begin{equation}
\label{eq:regSensors}
\begin{split}
\cD (\cE,\bparams) &\doteq \{\nabla \cdot \flow_k\}_{k=1}^{\numEvents}, \\
\cA (\cE,\bparams) &\doteq \bigl\{|\det(\mJ(e_k))|\bigr\}_{k=1}^{\numEvents},
\end{split}
\end{equation}
where the flow $\flow \doteq \partial \Warp(\bx,t;\bparams) / \partial t$, and Jacobian of warp $\mJ(\bx,t;\bparams) \doteq \partial \Warp(\bx,t;\bparams) / \partial \bx$, are the space-time derivatives of $\Warp$.
Just like \eqref{eq:IWE}, $\cD$ and $\cA$ are used to create images of average divergence and area deformation per pixel.
Finally, \cite{Shiba22sensors} computes $\regularizer$ as the trimmed mean of such images.

\subsection{Proposed Motion-based Regularizer}
\label{sec:method:proposedReg}

Although \cite{Shiba22sensors} successfully mitigates overfitting, it comes at a computational cost.
The complexity of these regularizers is $O(\numEvents + \numPixels)$ because \eqref{eq:regSensors} depends linearly on the number of events $\numEvents$ and the resulting average images have $\numPixels$~pixels.
This extra complexity makes the whole pipeline more than twice slower than the original (unregularized) CMax framework, whose complexity is also $O(\numEvents + \numPixels)$ \cite{Gallego18cvpr}. 
Not only the computational complexity is a burden, but also the fact that \eqref{eq:regSensors} are measured relative to a single reference time.
For example, $\cA(\cE, \bparams)$ increases as $t_k$ increases, since it measures the area deformation \emph{from $t_k$ to $\tref=t_1$}.
This scaling problem is undesirable because 
($i$) events far from $\tref$ contribute more to $\regularizer$ than events closer to $\tref$, 
and ($ii$) this effect could be amplified depending on the temporal distribution of the events.

\begin{figure}[t]
\centering
{\includegraphics[clip,trim={0cm 11.5cm 21cm 0cm},width=0.65\linewidth]{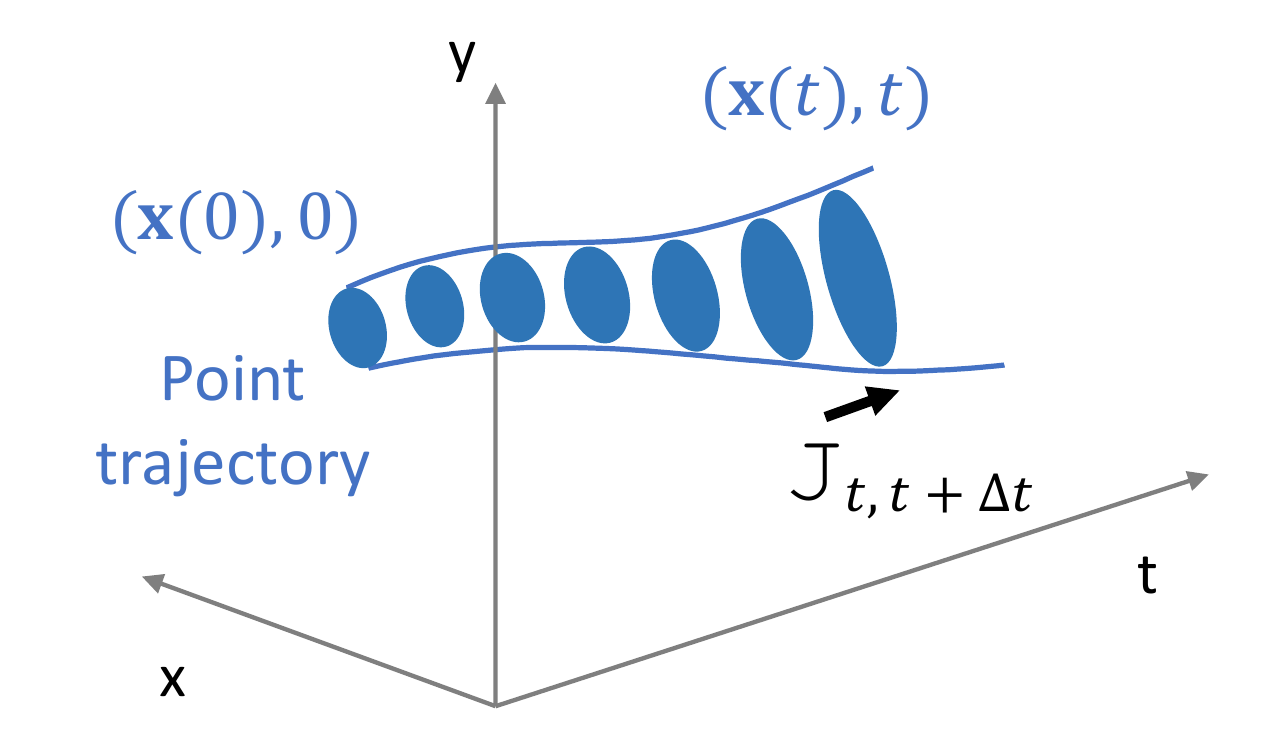}}
\caption{\emph{Rate of change of area deformation.} 
The warp $\Warp$ defines point trajectories $\bgamma(t)=(\bx(t),t)$ in the space-time image domain. 
We define the regularizer $\regularizer$ based on differential area deformation along $\bgamma(t)$. 
The rate of change of area is given by the derivative of the Jacobian $\mJdelta$.
}
\label{fig:method:absement}
\end{figure}

Intuitively, motion fields are well-posed or not (i.e., collapse-enabled) by design, regardless of the event data.
Hence, an ideal regularizer should not depend on the events, but solely on the warp parameters (\cref{fig:method:overview}, blue line).
The main idea of the proposed regularizer is to aggregate differential deformations rather than relative ones.
\Cref{fig:method:absement} shows the geometric interpretation: 
$\regularizer$ is obtained as the integral of the rate-of-change of the area element deformation along the space-time point trajectories $(\bx(t),t)$ defined by the motion.

\subsubsection{Collapse-enabled warp with 1~DOF}
\label{sec:method:oneDOF}

To illustrate our approach, consider the simplest example: a 1-DOF motion that approximates the translation of camera along its optical axis~$Z$.
This is a simplified zoom-in/out motion without knowledge of scene depth, as used in \cite{Mitrokhin18iros}.
The warp $\Warp$ is given by
\begin{equation}
\label{eq:warp:hz}
\bx'_k = (1 - t_k h_z)\, \bx_k,
\end{equation}
where $\bparams \equiv h_z$, and the coordinate frame is at the center of the image plane. %
For simplicity, $t \in [t_1, t_{\numEvents}]$ is normalized to~$[0,1]$.

Assuming an area element attached to each point of the motion trajectory $\bgamma(t)=(\bx(t),t)$ (\cref{fig:method:absement}), 
the change of area (i.e., area deformation) from $t$ to $t + \Delta t$ is given by:
\begin{equation}
\label{eq:deform:oneDOF}
|\mJdelta | \doteq \left| \det \biggl( \frac{d\bx(t + \Delta t)}{d\bx(t)} \biggr) \right|
= \biggl( \frac{1 - t h_z}{1 - (t + \Delta t) h_z} \biggr)^2.
\end{equation}

The Taylor series expansion of \eqref{eq:deform:oneDOF} at $\Delta t = 0$ is
\begin{equation}
\label{eq:taylor}
|\mJdelta | = 1 + \mJderivZero \Delta t + \cdots
\end{equation}
Since the first term is always 1 (i.e., is trivial), we focus on the second term, which conveys the meaning of ``speed'' of area deformation.
The derivative of \eqref{eq:deform:oneDOF} at $\Delta t = 0$ conveys the \emph{rate of change} or \emph{differential amplification factor} of the area:
\begin{equation}
\label{eq:rateofchange:oneDOF}
\mJderivZero = \frac{2 h_z}{1 - t h_z}.
\end{equation}

Finally, the total rate of change of the deformation along the observation time window is
\begin{equation}
\label{eq:regularizer:oneDOF}
\regularizer \doteq \int_0^1 \mJderivZero dt \stackrel{\eqref{eq:rateofchange:oneDOF}}{=} %
-2 \log |1 - h_z|.
\end{equation}

\begin{figure}[t]
  \centering
  \includegraphics[width=0.7\linewidth]{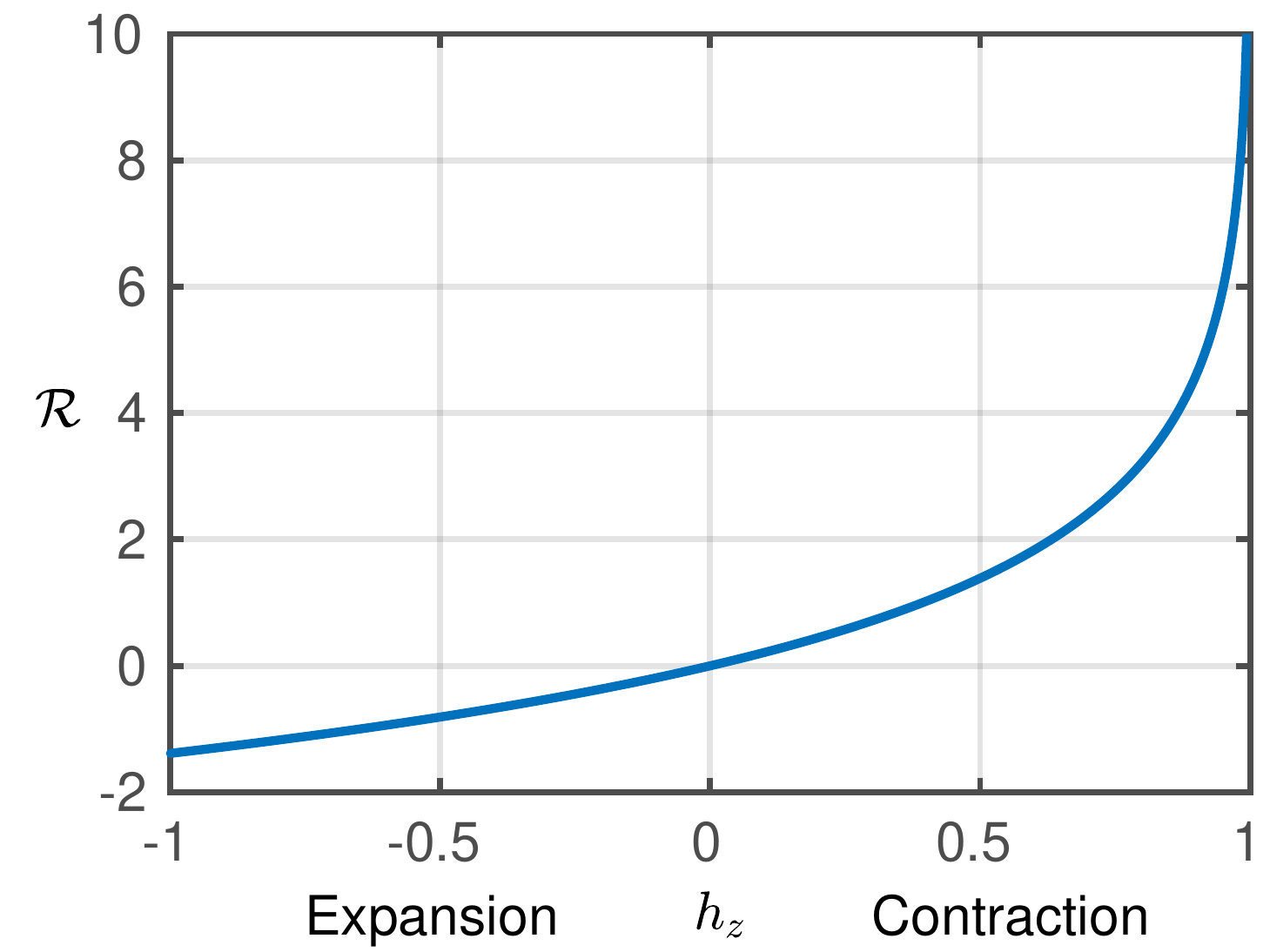}
\caption{Regularizer $\regularizer$ for the 1-DOF warp, \eqref{eq:regularizer:oneDOF}.}
\label{fig:regularizer:onedof}
\end{figure}

The regularizer \eqref{eq:regularizer:oneDOF} is plotted in \cref{fig:regularizer:onedof}. 
It solely depends on $\bparams\equiv h_z$ and has computational complexity $O(1)$.
In addition, it is developed from geometric principles, and it is interpretable:
$h_z = 0$ (identity warp) gives $\regularizer = 0$; 
$h_z \in (0,1)$ (contraction; collapsing warp) gives large $\regularizer > 0$; 
and $h_z < 0$ (expansion warp) gives $\regularizer < 0$.
Moreover, notice that $\regularizer$ behaves  like a barrier function, approaching infinity (i.e., large penalty) for values close to $h_z=1$ (maximum allowed contraction before events flip side with respect to the image center).

\subsubsection{Well-posed warp with 2 DOFs}
\label{sec:method:2dof}

The 2-DOF translational motion (feature flow) in image space is a well-posed warp, since collapse never happens because the motion lines are parallel.
The warp is given by $\Warp(\bx,t;\bparams) = \bx + (t -t_\text{ref}) \bparams$ with $\bparams \equiv (v_x, v_y)^{\top}$, assuming constant velocity $\bparams$ for all pixels.
Translations do not change the area element, i.e., $\mJdelta = 1$, 
hence $\regularizer \equiv \int_0^1 \frac{d|\mJ|}{d \Delta t} = 0$.
Since the resulting regularizer vanishes, it does not affect the landscape of the composite objective function \eqref{eq:compositeObjective}, as expected.

\subsubsection{Well-posed warp with 3 DOFs}
\label{sec:method:3dof}

Angular velocity estimation of a purely rotating event camera is a common research problem \cite{Gallego17ral,Peng21pami,Nunes21pami,Gu21iccv}.
This 3-DOF motion is another well-posed warp, although it involves small amounts of deformation \cite{Shiba22sensors}.
Assuming a small time interval, the warp is parametrized by the angular velocity $\bparams \equiv \angvel = (\omega_{x},\omega_{y},\omega_{z})^{\top}$. 
In calibrated homogeneous coordinates $\bx^h$, the point trajectories are described by \cite{Gallego18cvpr}:
\begin{equation}
\label{eq:warp:rotangvel}
\bx^{h} (t) \sim \Rot(\angvel t)\,\bx^h (0),
\end{equation}
where $t\in [0, 1]$ is normalized 
and $\Rot(\bphi) \doteq \exp(\bphi^\wedge)$ is parametrized using exponential coordinates \cite{Barfoot15book,Gallego14jmiv}.

The incremental rotation between $t$ and $t + \Delta t$ yields
\begin{equation}
\label{eq:warp:incrementalrot}
\bx^{h} (t + \Delta t) \sim \Rot(\angvel \Delta t)\,\bx^h (t).
\end{equation}

\begin{table*}[!t]
\centering
\caption{Results on the MVSEC dataset \cite{Zhu18rss}. 
The best values per column per group are in bold, and the second best are underlined. 
An asterisk in FWL indicates that event collapse occurred.
}
\adjustbox{max width=\textwidth}{%
\setlength{\tabcolsep}{2pt}
\begin{tabular}{ll*{10}{S[table-format=4.2]}}
\toprule
 &  & \multicolumn{5}{c}{Variance} & \multicolumn{5}{c}{Gradient Magnitude}
 \\
 \cmidrule(l{1mm}r{1mm}){3-7}
 \cmidrule(l{1mm}r{1mm}){8-12}
& 
&\text{AEE $\downarrow$} & \text{3PE $\downarrow$} & \text{10PE $\downarrow$} & \text{20PE $\downarrow$} & \text{FWL $\uparrow$}
&\text{AEE $\downarrow$} & \text{3PE $\downarrow$} & \text{10PE $\downarrow$} & \text{20PE $\downarrow$} & \text{FWL $\uparrow$}
\\
\midrule 
 & Ground truth flow & \novalue & \novalue & \novalue & \novalue & 1.047993099 & \novalue & \novalue & \novalue & \novalue & 1.047993099 \\
 & Identity warp & 4.845102657 & 60.58545033 & 10.38418053 & 0.314555551 & 1.0 &  4.845102657 & 60.58545033 & 10.38418053 & 0.314555551 & 1.0 \\
\midrule
\multirow{5}{*}{\begin{turn}{90}
1~DOF
\end{turn}} 
 & No regularizer & 89.33976 & 97.29911 & 95.41734707 & 92.39056633 & \anote{1.90} & 85.774515 & 93.96433 & 86.23523074 & 83.44562201 & \anote{1.87} \\
 & Whitening \cite{Nunes21pami} & 89.5848 & 97.18128 & 96.77280241 & 93.76030044 & \anote{1.90} & 81.10176919 & 90.86011026 & 89.0438589 & 86.20025493 & \anote{1.85} \\
 & Deformation \cite{Shiba22sensors} & 4.46584 & 52.60007 & 5.162230828 & \unumr{1.2}{0.127030408} & 1.0766 & 3.97013 & 48.7886 & \unumr{1.2}{3.209162316} & \unumr{1.2}{0.06638392314} & 1.0873 \\
 & Div. + Def. \cite{Shiba22sensors} & \unumr{1.2}{3.299381} & \bnum{33.08753545} & \bnum{2.613900583} & 0.4842627055 & \bnum{1.196874476} & \bnum{2.849103044} & \bnum{32.33705658} & \bnum{2.435868209} & \bnum{0.03246428393} & \bnum{1.174927461} \\
 & Ours & \bnum{3.16784037} & \unumr{2.2}{36.65428535} & \unumr{1.2}{4.010612447} & \bnum{0.09595173336} & \unumr{1.2}{1.155540142} & \unumr{1.2}{3.023115819} & \unumr{2.2}{34.46903954} & 3.39599112 & 0.07093179577 & \unumr{1.2}{1.165741409} \\

\midrule
\multirow{5}{*}{\begin{turn}{90}
4~DOF %
\end{turn}} 
 & No regularizer & 90.2153977 & 90.2153977 & 96.9406752 & 93.86453651 & \anote{2.05} & 91.26328556 & 99.48531996 & 95.06164699 & 91.46305309 & \anote{2.01} \\
 & Whitening \cite{Nunes21pami} & 90.81840915 & 99.1098107 & 98.03757781 & 95.0418796 & \anote{2.04} & 88.38464766 & 98.86976557 & 92.41199318 & 88.66373937 & \anote{2.00} \\
 & Deformation \cite{Shiba22sensors} & 8.132825707 & 87.4592456 & 18.53133953 & 1.092239517 & 1.027891481 & \unumr{1.2}{5.253162047}  & \unumr{2.2}{64.78740194} & \unumr{2.2}{13.17500467} & \unumr{1.2}{0.3739068138} & \unumr{1.2}{1.151436025} \\
 & Div. + Def. \cite{Shiba22sensors} & \unumr{1.2}{5.14482574} & \unumr{2.2}{65.60746589} & \unumr{2.2}{10.75469645} & \unumr{1.2}{0.3817697962} & \bnum{1.157302645} & 5.409725124 & 66.01420062 & 13.18880475 & 0.5417801136 & 1.14339947 \\
 & Ours & \bnum{4.359446773} & \bnum{58.63310397} & \bnum{6.556431157} & \bnum{0.1568843469} & \unumr{1.2}{1.148671166} & \bnum{4.300043269} & \bnum{54.26880905} & \bnum{5.612462037} & \bnum{0.3096051488} & \bnum{1.173120008} \\

\bottomrule
\end{tabular}
\label{tab:main_mvsec}
}
\end{table*}

\begin{table*}[!t]
\centering
\caption{Results on the DSEC dataset \cite{Gehrig21ral}. Same notation as \cref{tab:main_mvsec}.
}
\adjustbox{max width=\textwidth}{%
\setlength{\tabcolsep}{2pt}
\begin{tabular}{ll*{10}{S[table-format=4.2]}}
\toprule
 &  & \multicolumn{5}{c}{Variance} & \multicolumn{5}{c}{Gradient Magnitude}
 \\
 \cmidrule(l{1mm}r{1mm}){3-7}
 \cmidrule(l{1mm}r{1mm}){8-12}
& 
&\text{AEE $\downarrow$} & \text{3PE $\downarrow$} & \text{10PE $\downarrow$} & \text{20PE $\downarrow$} & \text{FWL $\uparrow$}
&\text{AEE $\downarrow$} & \text{3PE $\downarrow$} & \text{10PE $\downarrow$} & \text{20PE $\downarrow$} & \text{FWL $\uparrow$}
\\
\midrule 
 & Ground truth flow & \novalue & \novalue & \novalue & \novalue & 1.090988475 & \novalue & \novalue & \novalue & \novalue & 1.090988475 \\
 & Identity warp & 5.843080683 & 60.45375986 & 16.6455773 & 3.395154843 & 1.0 & 5.843080683 & 60.45375986 & 16.6455773 & 3.395154843 & 1.0 \\
\midrule
\multirow{5}{*}{\begin{turn}{90}
1~DOF
\end{turn}} 
 & No regularizer & 156.1258394 & 99.88354101 & 99.33309776 & 98.18140673 & \anote{2.58} & 156.0779055 & 99.92758934 & 99.40158847 & 98.10842233 & \anote{2.58} \\
 & Whitening \cite{Nunes21pami} & 156.1784653 & 99.95286001 & 99.51452798 & 98.26119902 & \anote{2.58} & 156.8175223 & 99.87981264 & 99.38352644 & 98.32937688 & \anote{2.58} \\
 & Deformation \cite{Shiba22sensors} & 9.007609466 & 68.95608551 & 18.86118538 & 4.773455209 & \bnum{1.402945911} & 5.793269017 & \unumr{2.2}{64.02245664} & 16.10840844 & \unumr{1.2}{2.751044423} & \bnum{1.361735211} \\
 & Div. + Def. \cite{Shiba22sensors} & \unumr{1.2}{6.061261761} & \unumr{2.2}{68.47937073} & \unumr{2.2}{17.08434678} & \bnum{2.272174046} & \unumr{1.2}{1.356512051} & \unumr{1.2}{5.526156883} & 64.09306651 & \unumr{2.2}{15.06454271} & \bnum{1.368051902} & \unumr{1.2}{1.348987971} \\ 
 & Ours & \bnum{5.805870209} & \bnum{57.18820089} & \bnum{14.7345445} & \unumr{1.2}{3.052300535} & 1.340454148 & \bnum{5.306213279} & \bnum{54.85118215} & \bnum{14.17229652} & 3.104782764 & 1.200084868 \\

\midrule
\multirow{5}{*}{\begin{turn}{90}
4~DOF %
\end{turn}} 
 & No regularizer & 157.542172 & 99.96821514 & 99.64288796 & 98.66698569 & \anote{2.64} & 157.3356017 & 99.94254934 & 99.52521317 & 98.44449633 & \anote{2.62} \\
 & Whitening \cite{Nunes21pami} & 157.7329839 & 99.9690844 & 99.66131966 & 98.71335243 & \anote{2.60} & 156.1180868 & 99.91218462 & 99.25821152 & 97.92947005 & \anote{2.61} \\
 & Deformation \cite{Shiba22sensors} & 15.11846767 & 94.95566648 & 62.58570184 & 22.6160355 & 1.24728993 & \unumr{2.2}{10.006316} & \unumr{2.2}{90.14615144} & \unumr{2.2}{39.45149429} & \unumr{1.2}{8.670040138} & \unumr{1.2}{1.249295937} \\
 & Div. + Def. \cite{Shiba22sensors} & \bnum{10.05902513} & \bnum{90.64687198} & \bnum{40.60743178} & \bnum{8.583359441} & \unumr{1.2}{1.263127072} & 10.39335383 & 91.01817547 & 41.81212268 & 9.396431802 & 1.226296549 \\
 & Ours & \unumr{2.2}{11.51437975} & \unumr{2.2}{91.50152207} & \unumr{2.2}{42.2883673} & \unumr{2.2}{11.05217135} & \bnum{1.296983665} & \bnum{9.554259428} & \bnum{88.94076493} & \bnum{35.95820572} & \bnum{7.742679675} & \bnum{1.309339262} \\
 
\bottomrule
\end{tabular}
\label{tab:main_dsec}
}
\end{table*}

Hence, the area element at $\bx(t)$ deforms according to:
\begin{equation}
\label{eq:deform:threeDOF}
\det \biggl( \frac{d\bx(t + \Delta t)}{d\bx(t)} \biggr) 
= \bigl( \br_{3}^\top (\angvel \Delta t) \bx^h(t) \bigr)^{-3},
\end{equation}
where $\br_{3}^\top$ is the third row of $\Rot(\angvel \Delta t)$ (see \cref{sec:append3dof}).
The derivative of \eqref{eq:deform:threeDOF} at $\Delta t=0$ is given by:
\begin{equation}
\label{eq:rateofchange3DOF}
\mJderivZero = 3 \bx^{h\top}(t) \angvel^\wedge \be_3.
\end{equation}

Finally, the integral of \eqref{eq:rateofchange3DOF} over the point trajectory (parametrized by the initial point $\bx(0)$) is given by:
\begin{equation}
\label{eq:regularizer:threeDOF}
\begin{split}
\regularizer_{\bx(0)} & = \int_0^1 \mJderivZero dt \\
&= 3 \omega_x \int_0^1 y(t) dt - 3 \omega_y \int_0^1 x(t) dt.
\end{split}
\end{equation}

The integrals in \eqref{eq:regularizer:threeDOF} have units of absement.
To obtain the regularizer~$\regularizer$, we threshold $\regularizer_{\bx(0)}$ at $-0.2$ and compute its mean,
which allows small amounts of natural deformation caused by rotation.
Similar to \eqref{eq:regularizer:oneDOF}, \eqref{eq:regularizer:threeDOF} does not depend on the events.
However, in contrast to \eqref{eq:regularizer:oneDOF}, \eqref{eq:regularizer:threeDOF} %
is spatially varying, providing an aggregated deformation map:
it is smaller in the center of the image and larger (in absolute value) in the periphery.
The computational complexity of $\regularizer$ is $O(\numPixels)$, which can be further reduced if only a subset of the pixels is used.

Although 3-DOF rotations involve small deformations, their values \eqref{eq:rateofchange3DOF} are considerably smaller than those of collapse-enabled warps like \eqref{eq:warp:hz}, and $\regularizer$ does not affect the accuracy of the angular velocity estimation (as \cref{sec:experim:rot} will show).
Also, pure rotations around the $Z$ axis $\angvel = (0, 0, \omega_z)^{\top}$ do not change the area, as expected, resulting in $\regularizer = 0$.

\subsubsection{Collapse-enabled warp with 4 DOFs}
\label{sec:method:4dof}

The 1-DOF warp (\cref{sec:method:oneDOF}) is a particular case of the 4-DOF warp in \cite{Mitrokhin18iros,Nunes21pami},
which approximates a freely-moving camera (6 DOFs) by means of a similarity transformation on the image plane.
The scaling parameter $h_z$ of the similarity transformation controls the amount of zoom in/out, i.e., the amount of contraction/expansion of the warp. 
Hence, we use \eqref{eq:regularizer:oneDOF} to penalize the amount of contraction. 
A mathematical justification is given in Appendix~\ref{sec:appendix}.
Since the purpose of our regularizer is to discourage the collapse while allowing a small amount of natural collapse, we use $\regularizer = \max \left(\alpha, -2 \log |1-h_z|\right)-\alpha$, with margin $\alpha=1$.

\section{Experiments}
\label{sec:experim}
\def\figWidth{0.1712\linewidth}
\def\figWidthLongA{0.21\linewidth} %
\def\figWidthLongB{0.22\linewidth} %
\begin{figure*}[ht!]
	\centering
    {\setlength{\tabcolsep}{1pt}
	\begin{tabular}{
	>{\centering\arraybackslash}m{0.35cm}
	>{\centering\arraybackslash}m{\figWidth} 
	>{\centering\arraybackslash}m{0.1cm}
	>{\centering\arraybackslash}m{\figWidth} 
	>{\centering\arraybackslash}m{\figWidthLongA}
	>{\centering\arraybackslash}m{\figWidthLongB}
	>{\centering\arraybackslash}m{0.1cm}
	>{\centering\arraybackslash}m{\figWidth}
	}
	 & \emph{Original}
 	 && \multicolumn{3}{c}{\emph{Without regularization}}
  	 && \emph{With regularization}\\
  	 \cmidrule(l{2mm}r{2mm}){2-2}
  	 \cmidrule(l{2mm}r{2mm}){4-6}
  	 \cmidrule(l{2mm}r{2mm}){8-8}
  	 
		{\rotatebox{90}{\makecell{MVSEC}}}
		&\gframe{\includegraphics[width=\linewidth]{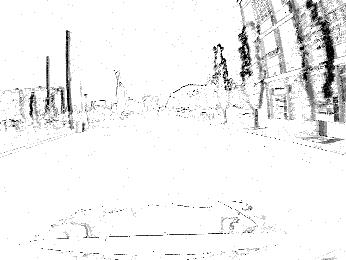}}
		&&\gframe{\includegraphics[width=\linewidth]{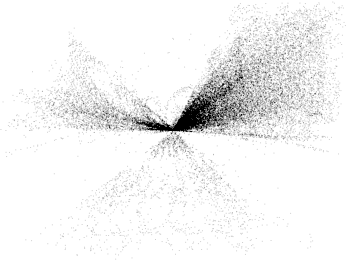}}
		&{\includegraphics[clip,trim={.3cm 0.8cm .3cm 0.8cm},width=\linewidth]{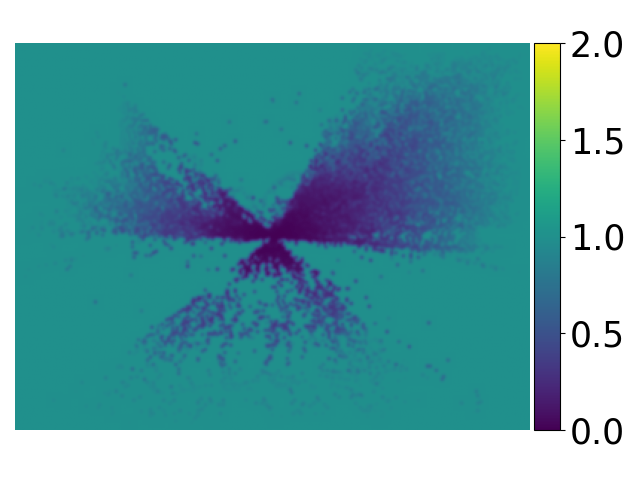}}
		&{\includegraphics[clip,trim={.3cm 1.1cm .3cm 0.9cm},width=0.97\linewidth]{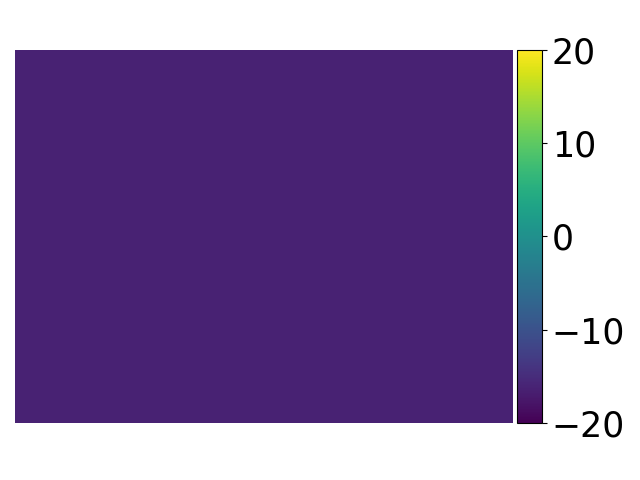}}
		&&\gframe{\includegraphics[width=\linewidth]{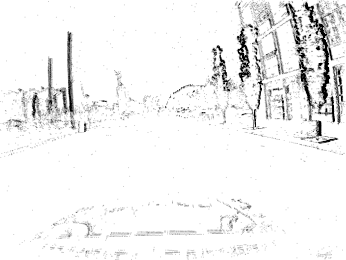}}
		\\

		{\rotatebox{90}{\makecell{DSEC}}}
		&\gframe{\includegraphics[width=\linewidth]{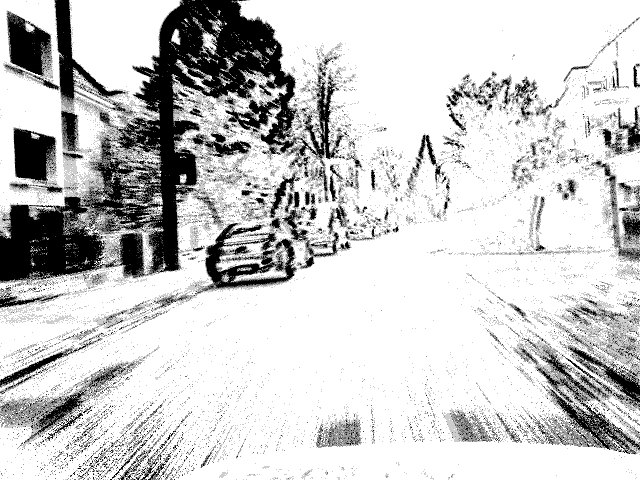}}
		&&\gframe{\includegraphics[width=\linewidth]{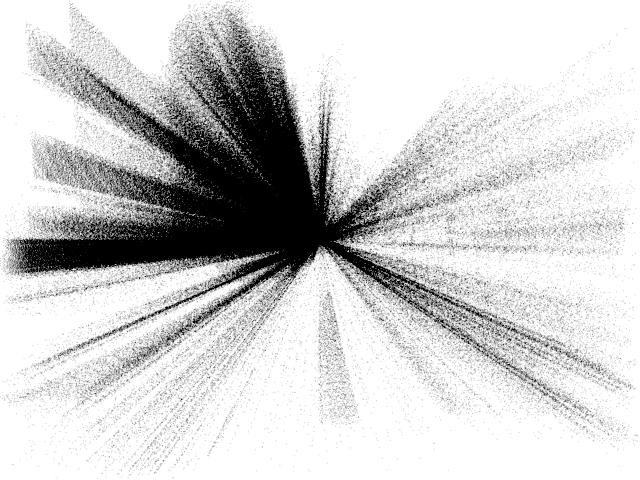}}
		&{\includegraphics[clip,trim={.3cm 0.8cm .3cm 0.8cm},width=\linewidth]{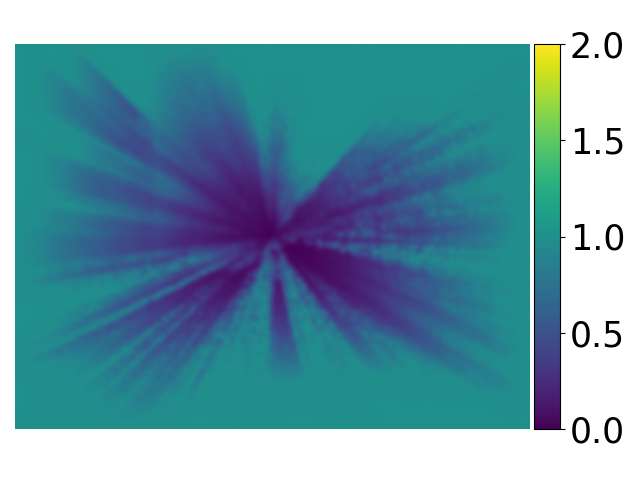}}
		&{\includegraphics[clip,trim={.3cm 1.1cm .3cm 0.9cm},width=0.97\linewidth]{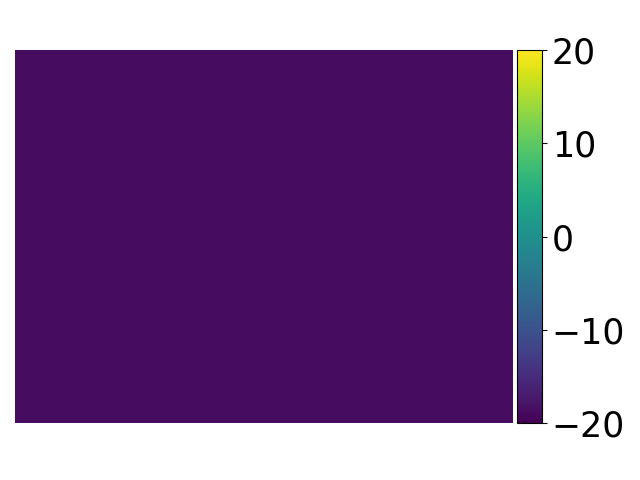}}
		&&\gframe{\includegraphics[width=\linewidth]{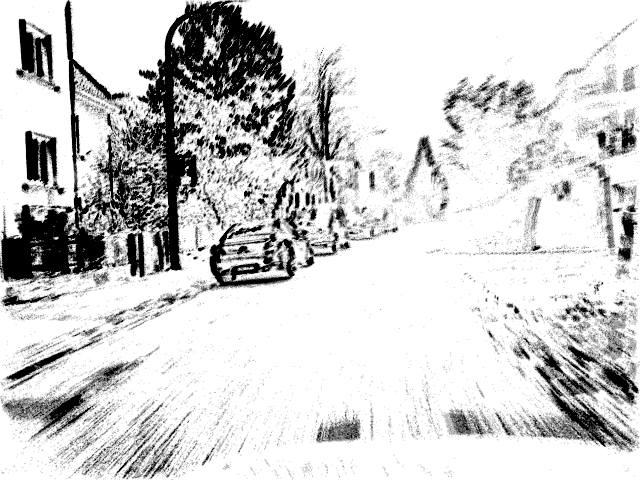}}
		\\

		{\rotatebox{90}{\makecell{boxes\_rot}}}
		&\gframe{\includegraphics[width=\linewidth]{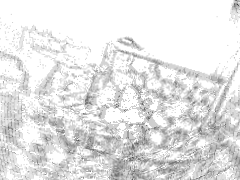}}
		&&\gframe{\includegraphics[width=\linewidth]{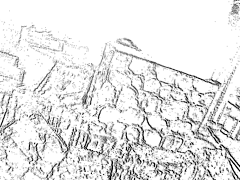}}
		&{\includegraphics[clip,trim={.3cm 1.1cm .3cm 1cm},width=\linewidth]{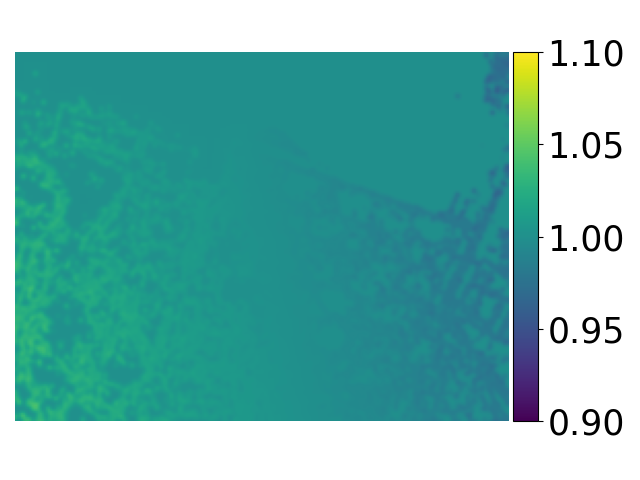}}
		&{\includegraphics[clip,trim={.3cm 1.1cm .3cm 1cm},width=\linewidth]{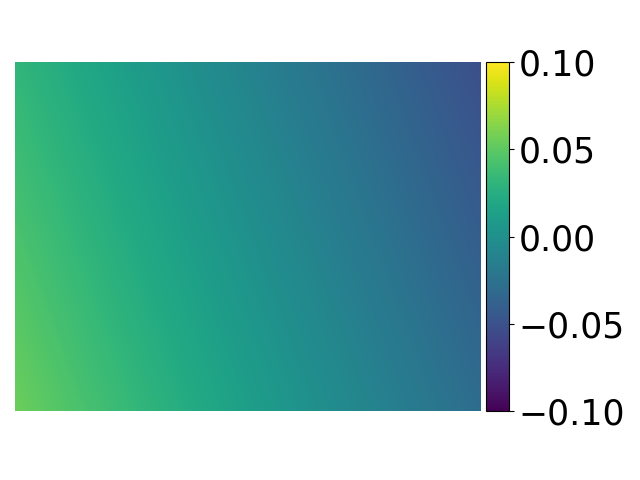}}
		&&\gframe{\includegraphics[width=\linewidth]{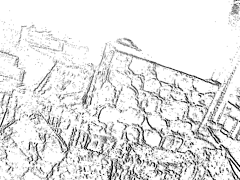}}
		\\

		{\rotatebox{90}{\makecell{dynamic\_rot}}}
		&\gframe{\includegraphics[width=\linewidth]{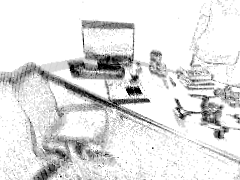}}
		&&\gframe{\includegraphics[width=\linewidth]{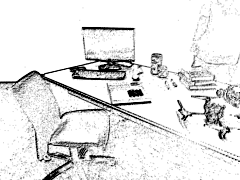}}
		&{\includegraphics[clip,trim={.3cm 1.1cm .3cm 1cm},width=\linewidth]{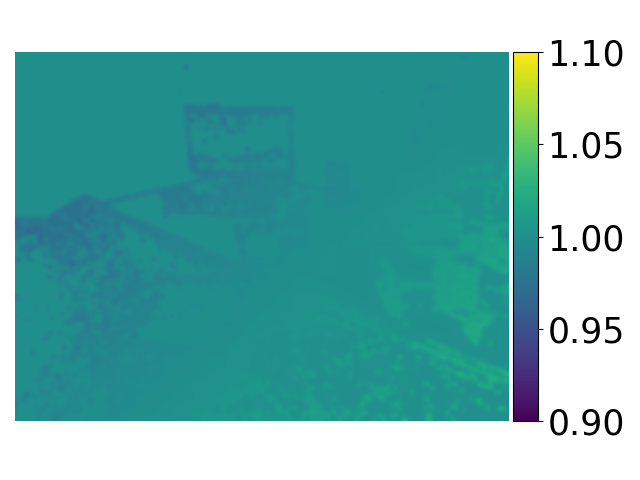}}
		&{\includegraphics[clip,trim={.3cm 1.1cm .3cm 1cm},width=\linewidth]{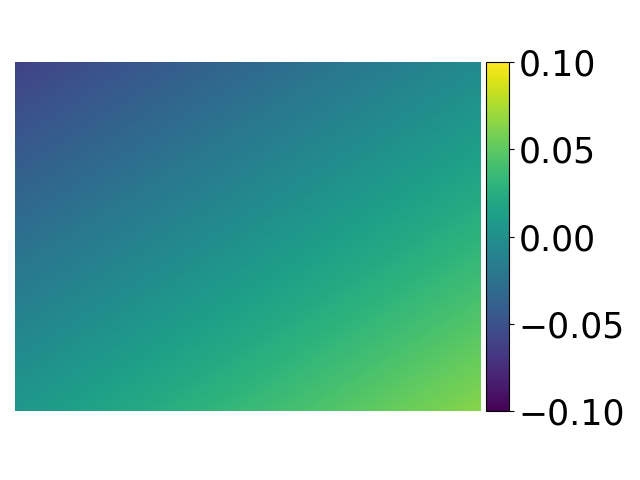}}
		&&\gframe{\includegraphics[width=\linewidth]{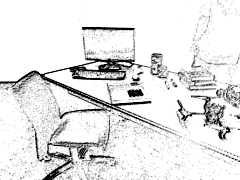}}
		\\

		& (a) Input events
		&& (b) Best IWE %
		& (c) Deformation \cite{Shiba22sensors}
		& \!\!\!\!\!\!\!\!\!\!\!  (d) Ours 
		&& (e) Best IWE %
		\\
	\end{tabular}
	}
	\caption{\emph{Qualitative results.} (a) Original events. (b)-(d) Results without regularization: 
	1-DOF motion results (MVSEC \cite{Zhu18ral} and DSEC \cite{Gehrig21ral}) are trapped in global optima of event collapse, as shown in the IWEs (b).
	The regularizers in such collapse cases (c)-(d) are very large compared with the well-posed warp cases (boxes\_rot and dynamic\_rot rows).
	(e) Results with the proposed regularizer: it mitigates collapse for MVSEC and DSEC scenes while it does not harm the ECD scenes.
	Best viewed in the electronic version.
    }
	\label{fig:main_compare}
\end{figure*}

We assess the performance of our regularizer by first showing its effectiveness on collapse-enabled warps, 
which naturally appear in driving sequences.
Second, a runtime analysis proves that our proposal is faster than prior work.
Finally, we also demonstrate the effect on rotational sequences, in order to show that the proposed regularizer does not harm well-posed warps. 
There is no need to test feature flow since we have proved analytically that the regularizer vanishes for such a warp.

\subsection{Setup: Datasets and Evaluation Metrics}
\subsubsection{Datasets}
\label{sec:experim:datasets}

The \emph{MVSEC} dataset \cite{Zhu18ral} is a standard dataset to evaluate various tasks, such as optical flow estimation \cite{Zhu19cvpr,Gehrig21threedv,Paredes21neurips,Shiba22eccv}.
The dataset consists of events, grayscale frames and IMU data from an event camera (mDAVIS346, $346 \times 260$ pixels \cite{Taverni18tcsii}), camera poses, LiDAR data, and ground truth optical flow provided by \cite{Zhu18rss}.
We use the outdoor sequences, where the event camera is mounted on a car.
Following previous work \cite{Shiba22sensors}, we select several excerpts from the \emph{outdoor\_day1} sequence that have a dominant forward motion, which is reasonably well approximated by collapse-enabled warps such as 1~DOF and 4~DOF cases.
In total, we evaluate on 3.2 million events spanning 10 s.

The \emph{DSEC} dataset \cite{Gehrig21ral} is another recent dataset of driving sequences.
It includes more complex scenes (e.g., moving objects, higher dynamic range) with a higher resolution event camera (Prophesee Gen3, $640 \times 480$ pixels).
Ground truth optical flow is computed as the motion field, using the scene depth from a LiDAR \cite{Gehrig21threedv}.
In total, we evaluate on 380 million events spanning 40 s from the \emph{zurich\_city\_11} sequence.

The \emph{ECD} dataset \cite{Mueggler17ijrr} is widely used to assess camera ego-motion \cite{Gallego17ral,Zhu17cvpr,Rosinol18ral,Gu21iccv,Rebecq17ral,Mueggler18tro,Zhou20tro,Shiba22sensors}.
Each sequence provides events, frames, calibration information, and IMU data from a DAVIS240C camera ($240 \times 180$ pixels \cite{Brandli14ssc}), as well as ground truth camera poses from a motion capture system (at 200Hz).
We use \emph{boxes\_rotation} and \emph{dynamic\_rotation} sequences for 3-DOF rotational motion estimation, to consistently compare with previous work.
In total, we evaluate on 43 million events (10 s) of the box sequence, and on 15 million events (11 s) of the dynamic sequence.

\subsubsection{Metrics}
\label{sec:experim:metrics}

Optical flow accuracy (for MVSEC and DSEC experiments) is given by the Average Endpoint Error (AEE) 
and the percentage of pixels with AEE greater than $N$ pixels (``$N$PE''), for $N=\{3,10,20\}$.
They are calculated only in pixels with valid ground truth.
We also adopt the FWL metric \cite{Stoffregen20eccv}, which is defined as the relative variance of the IWE with respect to that of the identity warp.
The FWL measures the IWE sharpness: FWL~$<1$ means that the estimation is worse than the zero-flow baseline, while FWL~$>1$ implies that the resulting IWE is sharper than the baseline.

Rotational motion accuracy is assessed as the RMS error of angular velocity estimation, following previous works \cite{Gallego19cvpr,Nunes21pami,Gu21iccv}.
Angular velocity is assumed to be constant during the time window of events, and compared with the IMU's gyroscope value (ground truth) at the midpoint: $(t_1 + t_{\numEvents}) / 2$.
We also use the FWL metric to measure the IWE sharpness.

The estimation time window spans: $dt=4$ grayscale frames (at $\approx$ 45Hz) in MVSEC (standard for MVSEC benchmark), 500k events for DSEC, and 30k events for ECD dataset, respectively.
For runtime comparison we use a fixed number of events (30k for MVSEC and 500k for DSEC), because the runtime depends on the number of events (e.g., $O(\numEvents + \numPixels)$).
We set $\lambda$ in \eqref{eq:compositeObjective} as follows: 
if the data term is the IWE variance, $\lambda=1.0$ for MVSEC and ECD experiments, and $\lambda=5.0$ for DSEC experiments;
if the data term is the squared magnitude of the IWE gradient, $\lambda=0.2$ for MVSEC and $\lambda=1.0$ for DSEC experiments.
The optimization algorithm is the Tree-Structured Parzen Estimator (TPE) sampler \cite{Bergstra11nips}.

\subsection{Results on Collapse-Enabled Warps}
\label{sec:experim:main}

\Cref{tab:main_mvsec,tab:main_dsec} report the results of collapse-enabled warp experiments (1 and 4~DOFs) on MVSEC and DSEC, respectively. 
They report the flow AEE, $N$PE, and FWL for the two data terms: image variance and the squared magnitude of the IWE gradient (``Gradient Magnitude'').
Throughout the experiments, both metrics are considerably high in the results of the original CMax (``No regularizer'') and whitening \cite{Nunes21pami} methods. 
This indicates that the warp overfits to events (event collapse).
On the other hand, our regularizer produces better AEE values and moderately higher FWL than $1$.
These results clearly show that our regularizer successfully discourages event overfitting while producing sharper IWEs than the identity warp.
Our results are competitive with those by \cite{Shiba22sensors}, and we do not find significant accuracy differences between them.

Qualitative results are shown in \cref{fig:main_compare} (MVSEC and DSEC rows).
Our regularizer (last column) provides the best IWEs, which reveal the sharp edges of the scene, while the IWEs without regularizer produce event collapse (second column).
Although $\regularizer$ in \eqref{eq:regularizer:oneDOF} is a scalar, we visualize it as an image to compare it with the area deformation map from \cite{Shiba22sensors}.
Notice that the area deformation map \cite{Shiba22sensors} shows collapse \emph{only} at pixels with warped events,
while our regularizer provides dense maps (even in pixels with no events, corresponding to homogeneous brightness regions) because it is purely geometric, based on the motion parameters.

\subsection{Runtime Comparison}
\label{sec:experim:runtimes}

\Cref{tab:runtime} reports the runtime comparison of the methods, notably with respect to the original CMax (``No regularizer'').
We use Python (3.9.12) on a CPU (Mac M1 2020, 8 Cores), and average the runtime over 400 trials.
The whitening technique \cite{Nunes21pami} is slower than the original CMax (``No regularizer''). 
The runtime difference is due to an extra SVD step on the events, which is more noticeable ($2\times$ slower) in the DSEC dataset than in MVSEC because it uses more events.
The ``Deformation'' regularizer in \cite{Shiba22sensors} is also two to three times slower than the original CMax.
When both regularizers in \cite{Shiba22sensors} are combined (``Div. + Def.''), the runtime becomes even larger.
Finally, our regularized approach has almost the same runtime as the original CMax, since its complexity is $O(1)$, thus it is two to four times faster than competing methods.

\begin{figure}[t]
\centering
{\includegraphics[clip,trim={6.6cm 1.9cm 8.5cm 3.5cm},width=0.8\linewidth]{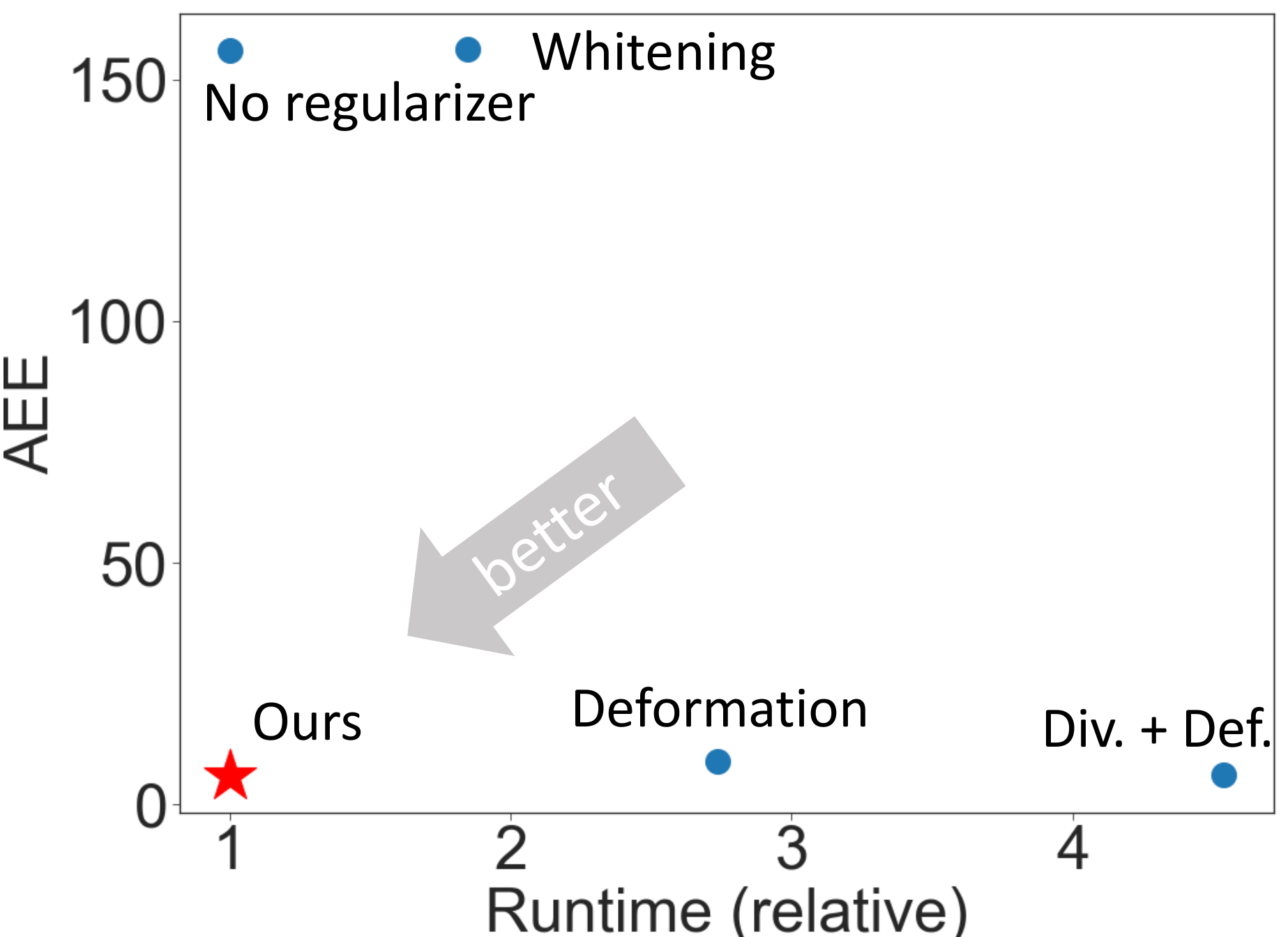}}
\caption{\emph{Runtime comparison} for the DSEC experiment. 
Runtime is relative to that of the original CMax (``No regularizer'').
Our method has desirable properties: small AEE and runtime.
}
\label{fig:runtime}
\end{figure}

\Cref{fig:runtime} visualizes the accuracy and runtime of the methods (on DSEC data).
Runtime is reported relative to the ``No regularizer'' case.
It clearly shows that the proposed regularizer is the only effective approach against event collapse that does not compromise the speed of the CMax framework.

\sisetup{round-mode=places,round-precision=1}
\begin{table}[!t]
\centering
\caption{Comparison of runtime (in milliseconds), averaged over 400 trials.
MVSEC: 30k events. 
DSEC: 500k events.
}
\adjustbox{max width=\textwidth}{%
\setlength{\tabcolsep}{4pt}
\begin{tabular}{l*{6}{S[table-format=3.1]}}
\toprule
 & \multicolumn{2}{c}{MVSEC} & &\multicolumn{2}{c}{DSEC} \\
 \cmidrule(l{1mm}r{1mm}){2-3} \cmidrule(l{1mm}r{1mm}){5-6}
& \text{Var.} & \text{Grad.} & &\text{Var.} & \text{Grad.}
\\
\midrule 
 No regularizer & \bnum{7.27} & \bnum{7.91} & &\bnum{111.285} & \bnum{112.8775} \\
 Whitening \cite{Nunes21pami} & 8.225 & 8.36 & & 205.4625 & 206.89 \\
 Deformation \cite{Shiba22sensors} & 20.22 & 21.1375  & & 304.44 & 307.6225  \\
 Div. + Def. \cite{Shiba22sensors} & 32.4075 & 31.6225 & & 505.02 & 506.1425 \\
 Ours & \unumr{1.1}{7.36} & \unumr{1.1}{8.025} & & \unumr{3.1}{111.4} & \unumr{3.1}{113.51}\\
\bottomrule
\end{tabular}
\label{tab:runtime}
}
\end{table}

\subsection{Results on Well-Posed Warps}
\label{sec:experim:rot}

To confirm that the proposed regularizer does not harm well-posed warps (e.g., 3-DOF rotational motion), we report results on the ECD dataset in \cref{tab:main_ecd}.
We use the variance as data fidelity and the Adam optimizer.
In both rotational sequences, the results of ``No regularizer'' and ours produce very similar RMS and FWL values.
This is because $\regularizer$ values are small and do not affect the landscape of the objective function.
As qualitatively shown in \cref{fig:main_compare}, 
for the 3~DOF experiments 
the absolute values of $\regularizer_{\bx(0)}$ are approximately less than $0.1$, 
while for the collapse-enabled warps they are larger than $10$ (i.e., two orders of magnitude difference).

\begin{table}[ht!]
\centering
\caption{Results on ECD dataset \cite{Mueggler17ijrr}.
}
\sisetup{round-mode=places,round-precision=3}
\label{tab:main_ecd}
\adjustbox{max width=\textwidth}{%
\setlength{\tabcolsep}{4pt}
\begin{tabular}{l*{4}{S[table-format=2.3]}}
\toprule
 & \multicolumn{2}{c}{boxes\_rot}
 & \multicolumn{2}{c}{dynamic\_rot}
 \\
 \cmidrule(l{1mm}r{1mm}){2-3}
 \cmidrule(l{1mm}r{1mm}){4-5}
&\text{RMS $\downarrow$} & \text{FWL $\uparrow$}
&\text{RMS $\downarrow$} & \text{FWL $\uparrow$}
\\
\midrule 
Ground truth pose & \novalue & 1.55901691 & \novalue & 1.414303624 \\
No regularizer & 8.8578556 & 1.561590243 & 4.823395392  & 1.420484652  \\
Deformation \cite{Shiba22sensors} & 8.664275512 & 1.560753596 & 4.822492508  & 1.4204841  \\
Div. + Def. \cite{Shiba22sensors} & 6.885339903 & 1.561584869 & 4.8219542  & 1.420483918  \\
Ours & 6.877290364 & 1.561589855 & 4.821937829 & 1.42048271 \\
\bottomrule
\end{tabular}
}
\end{table}

\subsection{Application: Time-to-contact}
\label{sec:ttc}
The parametrization of collapse-enabled warps has useful implications toward future application on intelligent vehicles, such as advanced driver-assistance system (ADAS).
Let us introduce another interpretation of the parameter $h_z$.
For a freely moving camera with linear and angular velocities $\linvel$ and $\angvel$, respectively, the apparent velocity $\velflow(\bx)$ on the image plane of a 3D point $\bX=(x,y,\depthx)^\top$ (at depth $\depthx$ with respect to the camera) 
can be computed using the $2\times 6$ feature sensitivity matrix \cite{Corke17book}:
\begin{equation}
\label{eq:featureSensitivity}
\velflow(\bx) =
\begingroup %
\setlength\arraycolsep{3pt} %
\begin{pmatrix}
\frac{-1}{\depthx} & 0 & \frac{x}{\depthx} & xy & -(1+x^2) & y \\[1ex]
0 & \frac{-1}{\depthx} & \frac{y}{\depthx} & 1+y^2 & -xy & x \\
\end{pmatrix}
\endgroup
\!\!
\begin{pmatrix}
\linvel\\[1ex]
\angvel\\
\end{pmatrix},
\end{equation}
which can be used to warp events: 
\begin{equation}
\label{eq:warp:flowhz}
\bx_k' = \bx_k - \velflow(\bx) t_k.
\end{equation}

Assuming a vehicle with body-frame velocity $v_z$, 
i.e., $\linvel \equiv (0, 0, v_z)^{\top}$, $\angvel \equiv (0, 0, 0)^{\top}$, 
the motion field \eqref{eq:featureSensitivity} becomes $\velflow(\bx) = (v_z / \depthx) \,\bx$,
and substituting in \eqref{eq:warp:flowhz} gives 
$\bx_k' = (1 - v_z/\depthx ) t_k.$
Comparing this expression to \eqref{eq:warp:hz}, 
and assuming $\depthx$ is spatially invariant, we identify
\begin{equation}
\label{eq:ttc}
h_z = \frac{v_z}{Z},
\end{equation}
i.e., the parameter $h_z$ is inverse of the time-to-contact or time-to-collision (TTC) \cite{Clady14fns}. %

\begin{figure}[t]
\centering
{\includegraphics[clip,trim={10.5cm 7.2cm 10.5cm 7.2cm},width=\linewidth]{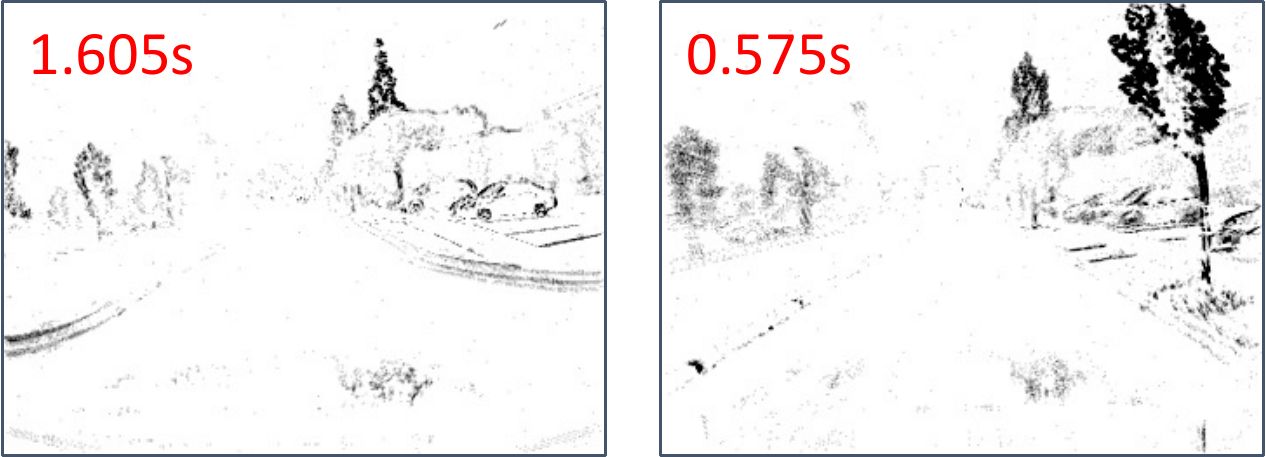}}
\caption{\emph{Time to Contact} application example.
The parametrization with $h_z$ in the 1-DOF warp can be used to approximate the TTC for the dominant depth of the scene represented by the events (e.g., the trees).
}
\label{fig:ttc}
\end{figure}

\Cref{fig:ttc} shows two examples of TTC from the MVSEC dataset.
It is remarkable that this 1-DOF warp model can be related to the popular concept in ADAS,
and our regularizer plays an important role toward real-time computation of TTC given its runtime.
Also note that \eqref{eq:ttc} establishes a relation between TTC, vehicle speed and scene depth, and that the TTC can be used to estimate the scene depth given the vehicle speed, 
or vice versa, the vehicle speed given the scene depth.
We hope this connection helps future implementation of event-camera application in collision avoidance systems.

\section{Limitations}
\label{sec:limitations}

As in many regularized problems, the regularizer weight $\lambda$ is empirically set. 
It depends on the paired objective function and on the data, i.e., on the scene. 
The proposed method also has a heuristically selected threshold (margin) that allows us to account for small natural deformations of the motion field. 
It would be desirable to develop automatic strategies to balance both data fidelity and regularizing terms (including the threshold) for best optimization convergence and results.

While we have obtained formulas to measure the rate of change of area deformation along point trajectories for low-DOF motions (up to 8-DOF homographies in Appendix~\ref{sec:appendix}), 
the regularizer requires aggregation in the form of integrals (e.g., \eqref{eq:regularizer:threeDOF}),
which are evaluated via numerical integration. 
Extending the ideas in this work to higher DOFs (e.g., optical flow) and developing efficient numerical approaches is an important research direction.

Our current implementation of the regularized CMax framework is not real time. 
This is especially relevant as the spatial resolution of event cameras increases (VGA size \cite{Gehrig21ral} and 1Mpixel event cameras \cite{Finateu20isscc}), which also increases the number of events to be processed. 
Therefore, in the future it will be important to speed up the method (both the data fidelity term and the regularizer) to enable interactive applications. 

Finally, while in the time-to-contact application (\cref{sec:ttc}) we established the connection between scene depth and vehicle speed, the example warp is 1-DOF (assumes a single depth for the whole scene), which may be an over-simplification.
The warp does not consider where events happen from a user perspective (e.g., on the road or on the side of the road). 
A more elaborate model would consider the location of the events and decide whether there will be an actual contact along the predicted vehicle trajectory.

\section{Conclusion}
\label{sec:conclusion}

We proposed a novel regularizer to mitigate event collapse in the CMax framework, based on aggregating differential deformations of the motion field.
The experimental results show its efficacy, achieving on-par state-of-the-art accuracy in low-DOF motion estimation problems.
Furthermore, the proposed regularizer is the only effective approach to date against event collapse that does not compromise the runtime of the CMax framework.
Since the low-DOF motion estimation forms a foundation of more complex motion models,
it would be important to analyze more complex warps, such as dense optical flow, to alleviate event collapse effectively and efficiently.
We hope this work encourages future research towards this paramount application of the CMax framework that leverages the advantages of event cameras.

\appendices
\section{}
\label{sec:appendix}

In homogeneous coordinates, a homographic warp $\Warp$ is given by \cite{Gallego18cvpr}
\begin{equation}
\label{eq:homographic:warp}
\bx^{h\prime}_k \sim \mH^{-1}(t_k; \bparams)\,\bx^h_k,
\end{equation}
and the point trajectories are represented by
\begin{equation}
\label{eq:homographic:motion}
\bx^{h} (t) \sim \mH(t; \bparams)\,\bx^h (0).
\end{equation}
Hence, the differential transformation from $t$ to $t+\Delta t$ is given also by a homography $\mHdelta$:
\begin{equation}
\label{eq:homographic:increment}
\bx^{h} (t+\Delta t) \sim \underbrace{\mH(t+\Delta t; \bparams)\, \mH^{-1}(t; \bparams)}_{\mHdelta} \bx^h (t).
\end{equation}

Using Result 1 in Appendix A of \cite{Rebecq18ijcv}, the determinant of the Jacobian $\mJ$ of the transformation (from $t$ to $t+\Delta t$) in Euclidean coordinates is
\begin{equation}
\label{eq:homography:detJ}
\det\bigl(\mJdelta\bigr) = \frac{\det( \mHdelta )}{(\be_3^\top \mHdelta \,\bx^h(t))^3},
\end{equation}
where $\be_3=(0,0,1)^\top$, $\bx^h(t)=(x(t),y(t),1)^\top$.

The 8-DOF motion admits several particular cases.

\subsection{Rotational Camera Motion}
\label{sec:append3dof}

Here, $\mH(t;\bparams) \equiv \Rot(t\angvel)$ in \eqref{eq:homographic:motion} is a $3\times 3$ rotation matrix. 
Rotation matrices have unit determinant and simplify: 
$\mHdelta %
= \Rot((t+\Delta t)\angvel) \, \Rot^{-1}(t \angvel)
= \exp\bigl(((t+\Delta t)\angvel)^\wedge\bigr) \exp\bigl((-t \angvel)^\wedge\bigr)
= \exp((\Delta t\, \angvel)^\wedge)
= \Rot(\angvel \Delta t)$.
This holds because the rotation axis $\angvel$ is unique.
Hence \eqref{eq:homography:detJ} becomes:
\begin{equation}
    \det\bigl(\mJdelta\bigr) = \frac1{(\be_3^\top \Rot(\angvel \Delta t) \,\bx^h)^3},
\end{equation}
which is \eqref{eq:deform:threeDOF}.
Computing the derivative yields \eqref{eq:rateofchange3DOF}:
\begin{align}
\mJderivZero &= \frac{-3 \bx^{h\top}(t)}{\bigl( \br_{3}^\top (\angvel \Delta t) \bx^h(t) \bigr)^{4}} 
\, \frac{d}{d \Delta t} \br_3 (\angvel \Delta t) \bigg|_{\Delta t = 0} \nonumber \\[0.8ex]
&= 3 \bx^{h\top}(t) \angvel^\wedge \be_3,
\end{align}
because
\begin{equation}
\begin{split}
\frac{d}{d \Delta t} \br_3 (\angvel \Delta t) &= \frac{d}{d \Delta t} \Rot^{\top} (\angvel \Delta t) \be_3 \\
&\approx \frac{d}{d \Delta t} \bigl( \mId - (\angvel \Delta t)^\wedge \bigr) \be_3 \\
&= - \angvel^\wedge \be_3 \\
&= (-\omega_y, \omega_x, 0)^{\top}.
\end{split}
\end{equation}

\subsection{Similarity Transformation on the Image Plane, $\displaystyle \operatorname {Sim}(2)$}

A planar similarity transformation has  4~DOFs and can be parametrized using linear image velocity, angular velocity around $Z$, and scaling speed on the image plane, $\btheta = (\velflow, \omega_z, s)^\top$ \cite{Nunes21pami}.
The point trajectories are given by \eqref{eq:homographic:motion} with 
\begin{equation}
\mH^S (t;\bparams) \doteq 
\begin{pmatrix}
\beta(t;s) \Rot(t \omega_z) & t \velflow\\
\bzero^\top & 1
\end{pmatrix}.
\end{equation}
Using \eqref{eq:homographic:increment} it gives
$\mHdelta = \mH^S (t+\Delta t; \bparams)\, (\mH^S)^{-1}(t; \bparams)$.
Similarities form a matrix Lie group, hence the inverse and the product of two similarities is also a similarity.
Since $\mHdelta$ is a similarity, its third row is $\be_3^\top \mHdelta = (0,0,1)$, which makes the denominator in \eqref{eq:homography:detJ} equal to one.
Substituting in \eqref{eq:homography:detJ} produces
\begin{equation}
\label{eq:deform:fourDOFsim}
\begin{split}
|\mJdelta | 
&= \left| \det \bigl( \mH^S (t+\Delta t; \bparams)\, (\mH^S)^{-1}(t; \bparams) \bigr) \right|\\[0.5ex]
&= \left| \frac{\det \bigl(\mH^S (t+\Delta t; \bparams) \bigr)}{\det \bigl( \mH^S (t; \bparams) \bigr)} \right|\\
&= \left(\frac{\beta(t+\Delta t;s)}{\beta(t;s)}\right)^2.
\end{split}
\end{equation}

Two intuitive remarks about \eqref{eq:deform:fourDOFsim}:
($i$) it only depends on the scaling DOF (i.e., one out of the four DOFs) since $\omega_z$ and $\velflow$ do not appear;
and ($ii$) it can be used to derive the 1-DOF formula \eqref{eq:deform:oneDOF}:
the 1-DOF scaling transformation is modeled by $\mH^S$ with $\omega_z = 0, \velflow=\bzero$ and $\beta(t; h_z) = (1-t h_z)^{-1}$. 
Substituting this choice of $\beta$ in \eqref{eq:deform:fourDOFsim} makes it coincide with \eqref{eq:deform:oneDOF}.

The 4-DOF transformation in \cite{Mitrokhin18iros} has a similar geometric meaning but a different parametrization.
Hence, we use the above result and penalize collapse by means of the corresponding scaling parameter in \cite{Mitrokhin18iros}.

\subsection{Affine Transformation on the Image Plane, $\displaystyle \operatorname {Aff}(2)$}

A planar affine transformation has 6 DOFs in $\bparams$. 
Letting
\begin{equation}
\mH^A (t;\bparams) \doteq 
\begin{pmatrix}
\mA(t;\bparams) & t \bvec \\
\bzero^\top & 1
\end{pmatrix},
\end{equation}
and using \eqref{eq:homographic:increment} gives
$\mHdelta = \mH^A (t+\Delta t; \bparams)\, (\mH^A)^{-1}(t; \bparams)$.
Affinities also form a matrix Lie group, hence $\mHdelta$ is an affinity. 
Moreover, its third row is also $\be_3^\top \mHdelta = (0,0,1)$, and
following similar steps as those in \eqref{eq:deform:fourDOFsim} yields
\begin{equation}
\label{eq:deform:affine}
\begin{split}
|\mJdelta | 
&= \left| \det \bigl( \mH^A (t+\Delta t; \bparams)\, (\mH^A)^{-1}(t; \bparams) \bigr) \right|\\[0.5ex]
&= \left| \frac{\det \bigl(\mA (t+\Delta t; \bparams) \bigr)}{\det \bigl( \mA (t; \bparams) \bigr)} \right|.
\end{split}
\end{equation}

Notice that the $2\times 2$ matrix $\mA$ includes not only a scaling parameter but also a shear, which affects the area deformation.

\section*{Acknowledgments}
This research was funded by the German Academic Exchange Service (DAAD), Research Grant-Bi-nationally Supervised Doctoral Degrees/Cotutelle, 2021/22 (57552338) and the Deutsche Forschungsgemeinschaft (DFG, German Research Foundation) under Germany’s Excellence Strategy – EXC 2002/1 ``Science of Intelligence'' – project number 390523135.

\balance
\bibliographystyle{IEEEtran}

\end{document}